%% file: elsarticle-template-num.tex
\documentclass[final,3p,times]{elsarticle}

\usepackage{amssymb}
\usepackage{amsmath}

\usepackage[utf8]{inputenc}
 \usepackage{bm}

\usepackage{amsthm}
\usepackage{amsmath}
\usepackage{parskip}
\usepackage{mathtools}
\usepackage{hyperref}
\usepackage[noabbrev,capitalise]{cleveref}
\usepackage{natbib}
\usepackage{geometry}
\usepackage[makeroom]{cancel}

\usepackage{lineno}
\usepackage{algpseudocode}
\usepackage{algorithm}

\usepackage{xcolor}
\usepackage{verbatim}

 \usepackage{csquotes}
 \usepackage{subcaption}
 \usepackage{bm}
\usepackage{mathrsfs}
\usepackage{graphicx}
\usepackage{amssymb}
\usepackage{smartdiagram}
\usepackage{xtab,siunitx,microtype}

\newcommand{\norm}[1]{\left\lVert#1\right\rVert}

\newcommand{\R}{\ensuremath{\mathbb{R}}}
\newcommand{\E}{\mathbb{E}}

\newcommand{\tit}[1]{\textit{#1}}
\crefname{appendix}{}
\linenumbers

\begin{document}

\begin{frontmatter}



\title{Gaussian Process Diffeomorphic Statistical Shape Modelling Out Performs Angle Based Methods for Assessment of Hip Dysplasia}


\author[a]{Allen Paul}
\ead{ap2746@bath.ac.uk}
\author[b]{George Grammatopoulos}
\ead{ggrammatopoulos@toh.ca}
\author[a]{Adwaye Rambojun}
\ead{adwayerambojun@gmail.com}
\author[e]{Neill D. F. Campbell}
\ead{nc537@bath.ac.uk}
\author[c]{Harinderjit S. Gill}
\ead{rg433@bath.ac.uk}
\author[a]{Tony Shardlow}
\ead{t.shardlow@bath.ac.uk}

\affiliation[a]{organization={Department of Mathematical Sciences},
  addressline={University of Bath}, 
  country={UK}}

\affiliation[b]{organization={The Ottawa Hospital},
  city={Ottawa},
  country={Canada}}

\affiliation[c]{organization={Department of Mechanical Engineering},
  addressline={University of Bath},
  country={UK}}
\affiliation[e]{organization={Department of Computer Science},
  addressline={University of Bath},
  country={UK}}


\begin{abstract}


Dysplasia is a recognised risk factor for osteoarthritis (OA) of the hip, early diagnosis of dysplasia is important to provide opportunities for surgical interventions aimed at reducing the risk of hip OA.  
We have developed a pipeline for semi-automated classification of dysplasia using $3D$ surface models obtained from volumetric CT scans of patients' hips and a minimal set of four clinically annotated landmarks on the acetabular rim (the most proximal, distal, anterior and posterior aspects), combining the framework of the Gaussian Process Latent Variable Model with diffeomorphism to create a statistical shape model, which we termed the  Gaussian Process Diffeomorphic Statistical Shape Model (GPDSSM). We used 192 CT scans, 100 for model training and 92 for testing. The GPDSSM effectively distinguishes dysplastic samples from controls while also highlighting regions of the underlying surface that show dysplastic variations. As well as improving classification accuracy compared to angle-based methods (AUC $96.2\%$ vs $91.2\%$), the GPDSSM can save time for clinicians by removing the need to manually measure angles and interpreting $2D$ scans for possible markers of dysplasia.
\end{abstract}



\end{frontmatter}



\input{intro}

\input{data}

\input{shapemodel}

\input{results}

\input{discussion}
\bibliographystyle{elsarticle-num} 
\bibliography{refs}

\input{Appendix}

\end{document}

%% file: intro.tex
\section{Introduction}
\begin{figure}[H]
\centering
    \includegraphics[width=.6\textwidth]{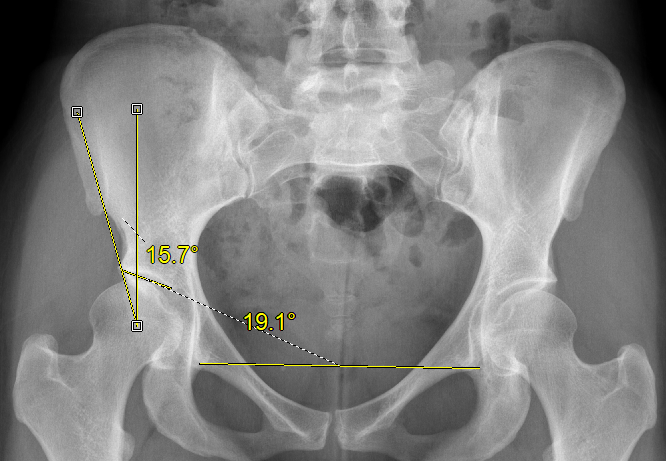}
       \caption{Example pelvic radiograph of a dysplastic subject, showing angle measurements, with LCEA angle of \ang{15.7} and AI of \ang{19.1}.}
    \label{Anglesfig}
\end{figure}

Osteoarthritis (OA) of the hip is one of the most common types of arthritis \cite{LongHuibin2022PToS} and leads to significant disability, negative impact on quality of life, and strain on healthcare systems \cite{ScheuingWilliamJ.2023Tboo}.  One of the main risk factors for developing hip OA is dysplasia of the hip \cite{ThomasG.E.R2014Sdot,WylesCodyC.2017TJCA}, a condition which is caused by abnormal development of the hip-joint region, resulting in a shallow acetabulum. 
(\cref{Anglesfig}). A shallow acetabulum leads to higher stress and adverse load distribution across the joint \cite{CheginiSalman2009Teoi}, causing damage to the acetabular cartilage and labrum, leading to hip osteoarthritis. The early diagnosis of dysplasia provides opportunities for surgical interventions that aim to reduce the future risk of hip OA. 

 
The current clinical methodology for detecting dysplasia is typically a radiographic or CT-based (2-dimensional) assessment of numerous angular and ratio measurements \cite{WilkinGeoffreyP.2017ACDo}. Amongst commonly used measures are the Lateral Centre Edge Angle (LCEA) and Acetabular Index (AI). For both measurements, there exist clinically agreed upon thresholds for LCEA $(<\ang{20})$ and AI ($>\ang{15}$) for identifying dysplasia, as well as thresholds for normality (LCEA$>\ang{25}$). Values of LCEA and AI between these ranges are considered borderline and need further evaluation. Given a CT scan of a $2D$ slice of the hip joint, clinicians may measure these angles using visualization software that is designed for radiological investigations. See \cref{Anglesfig} for how these angles are measured.
However, the measured angles on the scan can be sensitive to the patient's positioning during the scan, pelvic tilt, and the choice of $2D$ slice \cite{c8h_175976140,DeVriesZachary2021AMaS,VerhaegenJeroenC.F.2023ASAi}. Furthermore, this projection necessarily discards valuable information about the underlying $3D$ geometry of the acetabulum. Indeed, recent studies \cite{GraesserElizabethA.2023DtBD,IrieTohru2018Itaa} have indicated that ambiguous `borderline' dysplastic cases using the angle-classification system are better resolved by studying angles/ratios from $3D$ CT scans that take into account the full underlying geometry. Whilst CT scans are now commonly used for individuals presenting with symptoms consistent with dysplasia, it remains challenging and time-consuming for clinicians to manually measure multiple angle and ratio quantities on $3D$ scans, and often the data is treated like a $2D$ image.

\subsection{Existing Work}\label{existingworksec}
There have been multiple recent machine learning approaches \cite{NN1,NN2,NN3,NN4,NN5,NN6,NN7,NN8,NN9,NN10,NN11,NN12,NN13,NN15,NN16,NN17,NN18} in the dysplasia detection literature, aiming to detect dysplasia directly from scans. Many of these works rely on training CNN or UNET  architectures on raw $2D$/$3D$ CT and ultrasound scans, to classify/score patients for dysplasia. As with all machine learning approaches, these are very dependent upon the quality and quantity of available training data, in particular deep-learning approaches require substantial amounts of data, some authors quoting a minimum of 2300 training data sets \cite{Moon2020}. Much of the published reports utilize relatively small training-data sets. 

Another class of related methods to our work are statistical shape modelling (SSM) approaches \cite{SS1,SS2,SS3,SS4,SS5,SS6,SS7,SS8} focusing on either $2D$ landmark modelling of the hip joints, or $3D$ morphometric-modelling approaches, describing joint variation in the femur and acetabulum. As mentioned above, using $2D$ landmarks is susceptible to the orientation of radiographs. True $3D$ methods have the advantage of making use of the majority of the available data. 

\subsection{Aim}
Our aim was to create a novel semi-automated method based on shape modelling of the acetabulum to identify subjects with clinically relevant dysplasia. Currently, symptomatic individuals often risk being considered "normal" based on simple 2D assessments, but on detailed 3D assessments they are not. Thus, unless reviewed by experienced clinicians or have further detailed analysis, these individuals remain symptomatic and untreated for a long time. Improving diagnostic accuracy with new assessments would be very beneficial for this group, allowing them to be diagnosed and offered treatment earlier. The methods used were chosen as they are particularly suited to smaller data sets.

%% file: data.tex
\section{Methods}\label{datasec}




The data set for the current study consisted of $192$ volumetric CT scans of controls ($n=97$) and patients ($n=95$) with acetabular dysplasia. The CT data were collected at The Ottawa Hospital (TOH) and University College London Hospital (UCLH) \cite{c8h_175976140,DeVriesZachary2021AMaS,VerhaegenJeroenC.F.2023ASAi} as part of a study on risk factors of early onset osteoarthritis. The scans were fully anonymized. Each scan was accompanied by a sparse set of manually annotated landmarks on the acetabular ridges. The current study was approved by the Research Ethics Board of the Ottawa Health Science Network (OHSN-REB Protocol \#: 20220334-01H).

\subsection{Pre-processing to Extract Surface Data}\label{preprocsec}
The SSM that we used in this work (see \cref{Shape modelling}) is trained on surface data. As such, we pre-processed the CT data to extract the regions of interest of the pelvic surface, in particular the acetabular surface. 

\begin{figure}[H]
\centering
     \includegraphics[width=.28\textwidth]{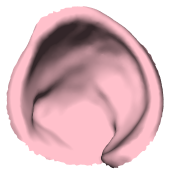}
     \hspace{50pt}
     \includegraphics[width=.3\textwidth]{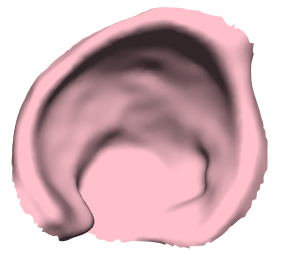}
         \\
        \includegraphics[width=.3\textwidth]{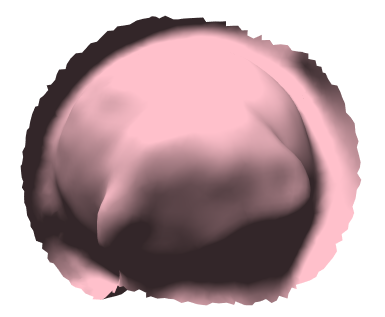}
          \hspace{50pt}
     \includegraphics[width=.3\textwidth]{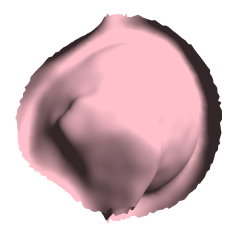}
     \\
     \includegraphics[width=.28\textwidth]{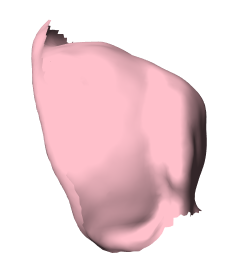}
       \hspace{50pt}
     \includegraphics[width=.28\textwidth]{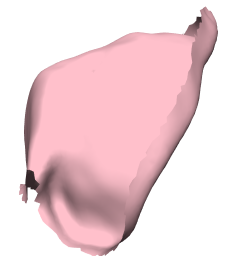}
     \caption{Example views of the left- and right- acetabular surfaces (in the corresponding columns), extracted from the hip surface of a dysplasia patient.}\label{extractionfig}
\end{figure}

The first step was to segment the pelvic region of the scan and extract surface data from the segmented scans to create a pelvic surface model. This was performed using the VTK toolkit \cite{vtkBook}, using a marching-cubes algorithm \cite{MarchingCubes}, designed to extract an isosurface from functional input data. For each scan in the dataset, the output of this algorithm produced a triangulated pelvic surface. The obtained surfaces were high resolution, with number of triangles $N\geq 10^{4}$.  
Next, we extracted the surfaces of the acetabuli. We used \cref{cup-extraction} for extracting this region in a semi-automated manner. This algorithm was applied separately to the left- and right-acetabular regions for each subject; see \cref{extractionfig}.
\begin{algorithm}[t]
	\caption{Acetabulum extraction algorithm}\label{cup-extraction}
	\begin{algorithmic}[1]
	\State Fit a plane of best fit to the annotated landmarks of the acetabular rim.
	\State Fit a rotation of the plane to the $(x,y)$ plane and apply this rotation to the acetabular region.
	\State Compute the minimal radius ball, which contains the rotated acetabulum, using ray casting.
	\State Extract all surface points within this ball above the $(x,y)$ plane.
 \State Remove stray disconnected components from extraction.
 \State The remaining surface is the acetabulum.
	\end{algorithmic}
	\end{algorithm} 
This method is semi-automated; it requires marking, on the pelvic surface model, of four points (the most proximal, distal, anterior and posterior aspects) on the acetabular rims for the first step of initial alignment. 
This information was already available in our dataset. This information is relatively fast to obtain for new scan data through manual annotation using visualization software. This step could be automated in future work. 

The final pre-processing step involved rigidly aligning all subjects' left and right acetabula. This removes all sources of variability between patients unrelated to the intrinsic geometric variation of the acetabular surface, namely rotation, scale and translation. This is a standard procedure in the pipeline for fitting a surface model. In the absence of point correspondences, we used a correspondence-less rigid-alignment method using the varifolds shape metric~\cite{Younes}. The varifold shape metric measures how different two shapes are by comparing both their geometric surfaces and the directions those surfaces face, making it ideal for quantifying differences between bone structures without needing to match specific points between them. Think of it as a sophisticated ruler that accounts for both the \enquote{bumpiness} and \enquote{tilt} of bone surfaces, giving you a single number that represents how much one bone shape deviates from another. All patient surfaces were aligned to a randomly chosen reference patient surface. The alignment algorithm is detailed in the Appendix.
We show exemplar post-alignment results in \Cref{alignedfig}.
\begin{figure}[H]
\centering
      \begin{subfigure}[t]{.9\textwidth}
      \centering
       \includegraphics[width=.35\textwidth]{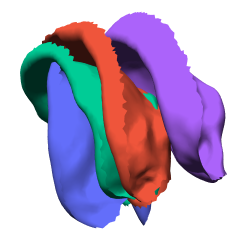}
              \hspace{50pt}
       \includegraphics[width=.3\textwidth]{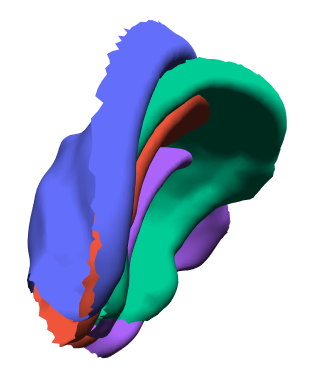}
       \\
       \centering
   \includegraphics[width=.35\textwidth]{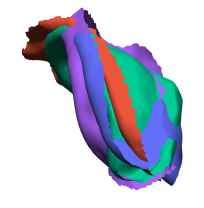}
          \hspace{50pt}
       \includegraphics[width=.35\textwidth]{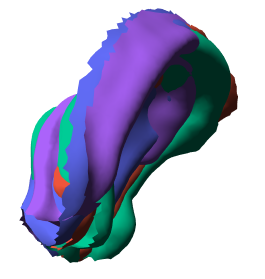}
       \end{subfigure}
     \caption{Pre-alignment (top row) and post-alignment (bottom row) for left- and right-acetabulum (in corresponding column), the the left and right surfaces were aligned individually using the green surface as the master surface for alignment.}\label{alignedfig}
\end{figure} 

The alignment process removed variation due to rotation, translation and scale. The remaining variation was due to natural population variation and disease mode variation. We made a simplifying assumption that these modes of variation are orthogonal. For each subject, we had the left- and right-acetabular surfaces. The spacing between the two surfaces was rescaled to be constant across all members of the data cohort. 



%% file: shapemodel.tex
\subsection{Latent Space and Diffeomorphism}
The concept of {\em latent space} is utilised in the present work. This concept originates in statistics and machine learning, and stems from classical concepts such as latent variable models. See \cite{bishop2007} for more on this topic. Given high-dimensional data, one models the data as the output of an unknown function mapping from a lower-dimensional \tit{latent} variable (typically a vector) to the high-dimensional data space. These lower-dimensional representations typically lie in a Euclidean space of vectors, which is referred to as the latent space. The resulting low-dimensional representation of data compresses the useful information about the original dataset into a much more compact representation. This concept may be seen as a generalization of traditional linear PCA-type models commonly used for dimensionality reduction. This idea has been applied to shapes many times before. See \cite{Younes} for background on this specific application.


Another concept we use repeatedly here is the notion of \textit{diffeomorphism}. For the purposes of this work, a diffeomorphism \cite{Younes} may be thought of as a smooth transformation deforming an anatomical object without causing self-intersections or tearing, preserving its overall structure. Modelling variation in shapes of anatomical objects may thus be reduced to modelling variations in diffeomorphisms that smoothly transform a reference anatomy. This is a well-established framework known as the LDDMM framework. See \cite{Younes} for a comprehensive reference.



\subsection{Shape Modelling}\label{Shape modelling}
Two Statistical Shape Models (SSMs) for the geometry of the acetabulum were constructed. These shape models automatically inferred low-dimensional latent representations of shape, effectively learning the underlying variation from the dataset of acetabuli. 
The first SSM used
the framework of Large Deformation Diffeomorphic Metric Mapping \cite{Younes} (LDDMM) theory. In particular, the geodesic shooting-based model of LDDMM \cite{Tangentstats}. This framework is widely used in computational anatomy \cite{Tangentstats,Younes} for modelling and understanding variations in anatomical shapes due to degenerative diseases. In LDDMM, shapes are modelled as deformations of a template shape under the action of a geodesic flow of diffeomorphism on the underlying domain. 

For the second SSM, we combined the probabilistic framework of the Gaussian Process Latent Variable Model (GPLVM) with the expressivity and topological guarantees of diffeomorphisms in order to yield a Gaussian Process Diffeomorphic Statistical Shape Model, correspondingly this SSM was termed the GPDSSM model.
In this model, each shape in the dataset was modelled as a noisy deformation of the template. Each deformation was generated as the flow of a random time-dependent vector field,  computed using an ODE solver. Fitting this model requires more care than the standard GPLVM, as our model was fit to triangulated surfaces without known correspondences, and so required geometric likelihoods and noise models. Further mathematical details are given in \ref{GPLVM background}. This GPDSSM may be seen as a nonlinear generalization of the LDDMM model, resulting in a more expressive parametrization.

The SSMs were fitted using the surface data from 100 CT scans, consisting of 43 scans of subjects with dysplasia and 57 scans of control subjects.

\subsection{Classification in Latent Space}
The latent space is typically a low-dimensional Euclidean space. During training, the shape model will adjust its parameters to best mimic the real-world distribution of shapes in the training data. In the compressed lower-dimensional latent-space representations learned by the model, similar shapes cluster together, whereas dissimilar shapes tend to lie further apart. The SSM models were fitted to the surfaces without access to the labels that describe whether the surfaces were from a control or dysplastic individual. Natural correlations exist between the shapes and the labels due to connections between the geometry of the acetabulum and the status of dysplasia. Shapes with similar dysplastic variations will cluster together in the low-dimensional latent-space embedding. This observation was exploited to train classification algorithms for each SSM based upon the correlations in the latent-space representation between shape and dysplastic status. These classifiers were trained using the labelled surface data from the same 100 CT scans used to fit the SSMs. A probabilistic classifier was chosen to encode uncertainty in the latent space classification, outputting a probability score in the range $[0,1]$ indicating the likelihood of the presence of dysplasia. 

Once the SSMs and classifiers were trained these were then used to determine the probability of dysplasia in the test data set, \Cref{flowchart} shows the overall process of model fitting and usage.


\begin{figure}[h]

\usesmartdiagramlibrary{additions}
\usetikzlibrary{arrows}
\begin{minipage}[t][3.5cm]{\textwidth}
\begin{center}
\smartdiagramset{
uniform color list=black!40!white for 3 items,
back arrow disabled=true,
 border color=gray,
uniform connection color=true,
additions={
additional item bottom color=black!40!white,
additional item offset=0.95cm,
additional item border color=gray,
additional arrow color=black!40!white,
additional arrow tip=to
}
}
\smartdiagramadd[flow diagram:horizontal]{%
Pre-process,Fit statistical shape model,Classify as dysplastic or healthy%
}{%
below of module1/CT training data, below of module3/New data%
}
\smartdiagramconnect{-to}{additional-module1/module1}
\smartdiagramconnect{-to}{additional-module2/module3}
\end{center}
\end{minipage}

    \caption{Flow chart showing process of fitting and use of the statistical shape model.}
    \label{flowchart}
\end{figure}

\subsection{Testing and Comparison with Angle Based Classification}
The testing and comparison were based on 92 data sets, which were not used for training of the SSMs and classifiers. These were from 40 control and 52 dysplasia subjects, for these subjects the current clinically accepted LCEA and AI angles were available. Angle-based classification was performed by combining the LCEA and AI angles, using LCEA $(<\ang{20})$ and AI ($>\ang{15}$) to indicate the presence of dysplasia. The comparison between the angle-based and two SSM-based classifers, namely LDDMM and GPDSSM, performance was made by plotting ROC curves and computing AUC scores.

\subsection{Cross-validation}
In order to formally test the hypothesis that our GPDSSM model outperformed the angle-based classification, we performed a leave-one-out cross-validation procedure (LOOCV) using a bootstrapping approach, resampled $100$ times over each fold to obtain a distribution of AUCs for both methods and estimate the test error. The LOOCV procedure also allowed the estimation of confidence intervals for accuracy, sensitivity and specificity.

 We then performed a Wilcoxon signed rank test on the obtained distributions of AUCs. The null hypothesis posits there is no difference in the median AUC, and the alternate that the AUC distribution of the GPDSSM is stochastically greater than the angle AUC distribution. The test was performed on the difference distribution with $\alpha=0.05$.

\subsection{Visualisation of surface variations}\label{visualisationsec}
It is possible to identify and visualise the key geometric variations of the acetabulum that the GPDSSM model uses to distinguish dysplastic patients from controls. Since GPDSSM parametrizes variations in shape as deformations of a fixed template, it is possible to further quantify the strength of local shape variations of a given patient's acetabulum. This property makes the GPDSSM model much more interpretable than existing methods for dysplasia classification that directly classify CT volumes.

Since all shapes in the shape dataset were modelled as deformations of a fixed template $T$, the class averages were obtained by averaging the landmarks of the template deformations within the controls and dysplastics subsets of the data. The averaging process minimized the effects of between-patient shape variations and highlighted the common modal shape variation within each class.

In order to quantitatively identify regions that show the greatest difference in shape between controls and dysplastics, the pointwise differences from the template and fitted shapes were computed in the dysplastics and controls classes respectively. Since the model automatically placed all shapes in correspondence, it was possible to compare the difference distributions pointwise, conducting multiple (BH-adjusted) permutation tests to assess the regions with statistically significant differences between controls and dysplastics. 

Visualisation of the principal modes of shape variation due to dysplasia encoded in the latent-space model gives further insight into acetabular shape. For each control shape in the dataset, we computed the closest dysplastic shape encoded in the GPDSSM latent space. Subsequently, we computed the minimal deformation (under the model) required for this control to progress to dysplastic status. This quantity minimises the effect of cross-patient variation, as it is computed across shapes that are already close in latent space.  The distribution of minimal deformations from controls to dysplastics was establised across the whole dataset. Principal components analysis (PCA) was then performed on the obtained residuals across all controls to obtain the modes of shape variation transforming controls to dysplastics. We applied the principal mode of these transformations to the generic template.


%% file: results.tex
\section{Results}\label{Numexpsec}


\subsection{GPDSSM}
\begin{figure}[H]
      \begin{subfigure}[b]{1\textwidth}
\centering
         \includegraphics[width=.8\textwidth]{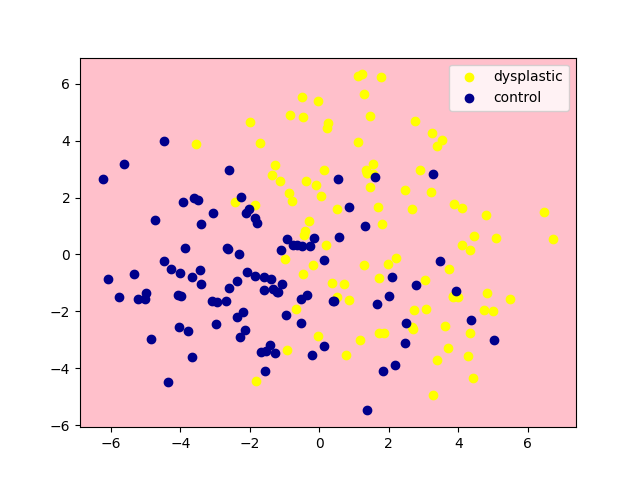}
           \end{subfigure}
     \caption{Two-dimensional visualization of training and test data latent embeddings for the GPDSSM, using PCA projection.}\label{latentspacevis}
\end{figure}

For the GPDSSM, the learned latent space was $20$-dimensional. To visualize this space, a two-dimensional plot was generated using a principal component analysis (PCA) projection, cross-plotting the two dominant principal components (\cref{latentspacevis}). 
We observed that, even without label information, the GPDSSM model learnt to automatically separate controls and dysplastics in the latent space. This is due to the subtle differences in shape between the two groups due to dysplastic surface variations. 

\subsection{Classification Performance}\label{classificationsec}
\begin{figure}[H]
      \begin{subfigure}[b]{1\textwidth}
    \centering
         \includegraphics[width=1.\textwidth]{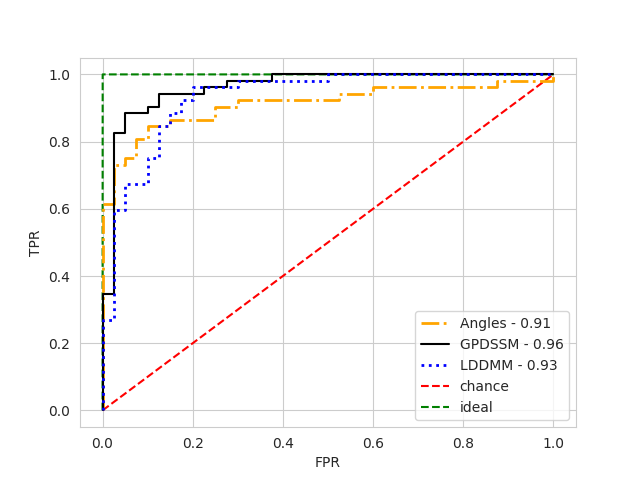}
    \end{subfigure}
     \caption{ROC curves and AUC scores estimated from LOOCV, for angle-based (Angles),the GPDSSM  and LDDMM classification methods.}\label{ROCfig}
\end{figure}

Classification for dysplasia performance comparisons between angle-based, LDDMM and GPDSSM are shown \cref{ROCfig} and in \cref{classificationtable}.
\Cref{classificationtable} shows a comparison of classification results in terms of overall estimated test accuracy (for a $50\%$ threshold) from LOOCV, estimated AUC, dysplastic class performance (Sensitivity), and control class performance (Specificity). The best-performing representation in each category is highlighted in bold.    
\begin{table}[H]
\centering
\scalebox{1.1}{
 \begin{tabular}{||c c c c c||} 
 \hline
 Model & AUC & Accuracy & Specificity & Sensitivity \\ [0.5ex] 
 \hline\hline
 Angle based & 0.91 [0.84, 0.97]  & 0.87 [0.79, 0.93] & \textbf{0.90 [0.8, 0.98]} & 0.85 [0.74, 0.94]  \\
 LDDMM & 0.93 [0.88,0.98] & 0.88 [0.79,0.93]  & 0.83 [0.70,0.93] & 0.90 [0.81, 0.97] \\
 GPDSSM & \textbf{0.96 [0.92, 0.99]}  & \textbf{0.90 [0.84, 0.96]} & 0.86 [0.73, 0.95] &  \textbf{0.94 [0.87, 1.0]} \\ [1ex] 
 \hline
 \end{tabular}}
 \caption{Comparison of classification, in terms of AUC, Accuracy, Specificity and Sensitivity. In square brackets are given $95\%$ bootstrap confidence intervals for each quantity.  } \label{classificationtable}
\end{table}
The GPDSSM outperforms the angle-based representation in terms of AUC ($96.2$\% to $91.2$\%), suggesting that the overall classification based on our latent-space representation is more discriminative across all classification thresholds in separating dysplastics from controls. The GPDSSM representation also yields an improved sensitivity (performance on dysplastic) over the angle-based representations ($94.2$\% to $84.6$\%), suggesting that this representation is particularly good at identifying dysplastic cases. In contrast, angle-based representations are more likely to incorrectly classify dysplastic as controls. This is an important area of improvement, as type II errors are far more costly in such a clinical setting, where patients may face significant deterioration of health if not treated due to false negatives.

In \cref{classificationtable}, the GPDSSM outperforms the standard LDDMM across all scores, which indicates the increased expressivity and powerful nonlinear dimensionality reduction afforded by our GPLVM-based diffeomorphic framework. We also remark that while the angle-based method has the best specificity (performance on controls) of $90$\%, it underperforms in terms of sensitivity against \textit{both} LDDMM ($92.3$\% to $84.6$\%) and GPDSSM ($94.2$\% to $84.6$\%). This suggests the use of latent-space representations based on the entire three-dimensional geometry has a measurable advantage in detecting dysplastic cases compared to the simpler angle-based representations. 

\subsection{LOOCV}

\begin{figure}[H]
\hspace{-25pt}
      \begin{subfigure}[b]{.55\textwidth}
         \includegraphics[width=\textwidth]{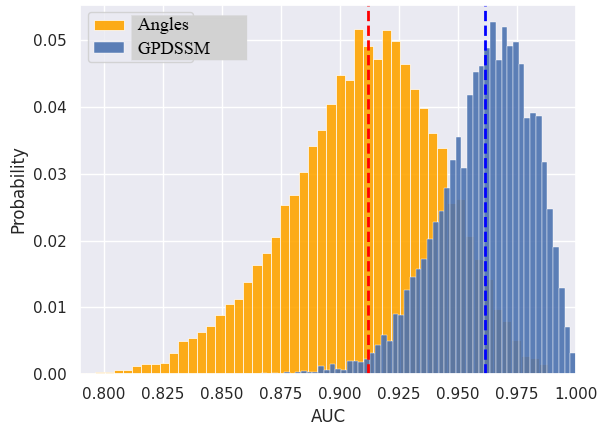}
    \end{subfigure}
    \begin{subfigure}[b]{.55\textwidth}
         \includegraphics[width=\textwidth]{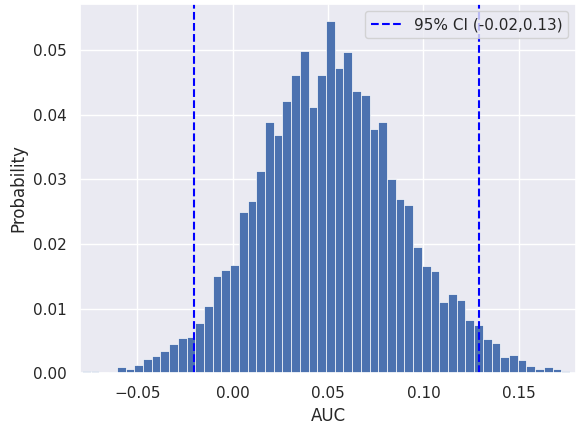}
    \end{subfigure}
    \caption{Distribution of AUCs obtained by angle-based (Angles) and GPDSSM methods (left), and difference distribution (right).}\label{Hists}
\end{figure}

\Cref{Hists} illustrates the distribution of AUCs for angle-based and GPDSSM methods and the paired differences across each bootstrap sample. 
The distribution of GPDSSM AUC scores visually suggests a difference of location to the angle-based method (\cref{Hists}.



\subsection{Visualisation of surface variations}

\begin{figure}[H]
\centering
\includegraphics[width=.25\textwidth]{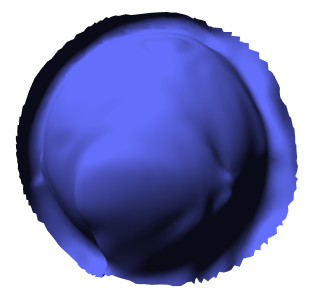}\includegraphics[width=.25\textwidth]{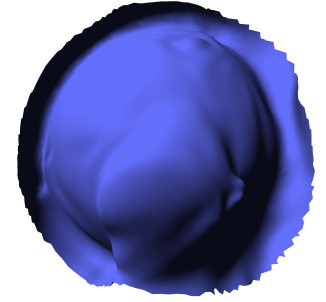}
     \caption{Average shapes for control (left colum) and dysplastics (right column). Top to bottom,  side profile, outer cup and inner cup views of the acetabulum respectively.}\label{averagesfig}
\end{figure}

The average acetabula surfaces for control and dysplasia are shown in \cref{averagesfig}.
\begin{figure}[H]
\centering
\includegraphics[width=.25\textwidth]{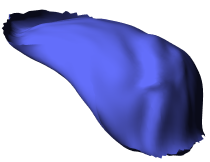}
\includegraphics[width=.25\textwidth]{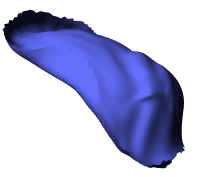}
\end{figure}
\begin{figure}[H]
\centering
\includegraphics[width=.25\textwidth]{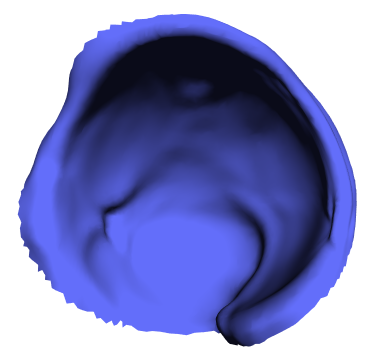}\includegraphics[width=.25\textwidth]{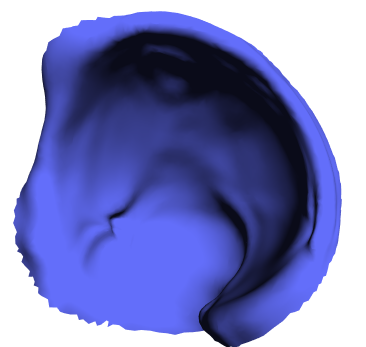}
\end{figure}

In \cref{averagesfig}, the average dysplastic acetabular shape has a much shallower cup structure compared to the control class, as well as reduced inner-cup volume, both of which are factors that influence the closeness of fit between the femoral head and acetabulum.

\begin{figure}[H]
\centering
\includegraphics[width=.33\textwidth]{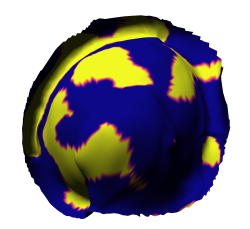}
     \caption{Visualisation of acetabular regions with $p$-adjusted statistically significant differences (at level $\alpha=0.05$) between controls and dysplastics highlighted in yellow. }\label{pvalfig}
\end{figure}

The regions with statistically significant differences between controls and dysplastics are shown in \cref{pvalfig}. 

In \cref{pvalfig}, the regions of statistically significant variations are visually consistent with the observed differences in mean control and dysplastic shapes in \cref{averagesfig}. In particular, the majority of the significant variations are concentrated on the cup boundary---associated with boundary retraction--- and the apex/top of the cup---associated with flattening of the top part of the cup and shrinking cup volume.

\begin{figure}[H]
\centering
\includegraphics[width=.4\textwidth]{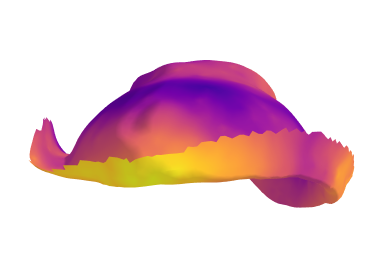}
\includegraphics[width=.4\textwidth]{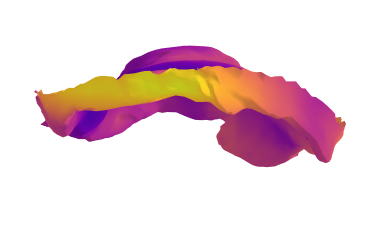}\\
\includegraphics[width=.4\textwidth]{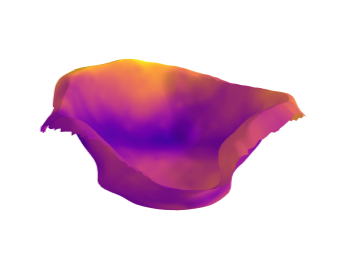}
\includegraphics[width=.4\textwidth]{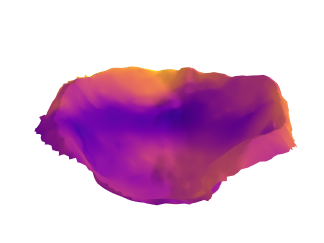}\\ 
 \caption{Variation along main PCA mode in differences between dysplastics and controls. (Left column): mean control shape of left acetabulum in two views. (Right column): corresponding shapes when top dysplastic PCA mode is applied.}\label{PCAFIG}
\end{figure}

The output of the PCA on the obtained residuals across all controls to obtain the modes of shape variation transforming controls to dysplastics applied to the generic template is visualised in \cref{PCAFIG}. The brighter regions indicate areas with greater deformation.
We observe from the heat maps of \cref{PCAFIG} that the modal deformation is predominantly focused on the acetabular boundaries and apex. Deformation in these regions drives the boundary to retract and gives the cup a shallow structure, as observed in both the mean and mode shape variations of \cref{PCAFIG}. We also observe inwards deformation on the apex of the cups, which leads to both flattening of the upper part of the acetabulum.

%% file: discussion.tex
\section{Discussion}
Dysplasia is a significant risk factor for the development of hip osteoarthritis 
\cite{Jacobsen2005}. There is growing evidence that surgical intervention with periacetabular osteotomy improves outcomes and reduces risk of hip OA \cite{Clohisy2009,Sohatee2020}. Diagnosing dysplasia can be challenging; the most common clinical methods rely on measurements made on radiographs or 2D CT projection. However, the measured angles are known to be sensitive to patient positioning, pelvic tilt, and the choice of $2D$ slice \cite{c8h_175976140,DeVriesZachary2021AMaS,VerhaegenJeroenC.F.2023ASAi}. There is a need for more robust methods for identifying individuals with dysplasia who will benefit from surgical intervention. 

In this work, we have developed a semi-automated algorithm to classify the presence of dysplasia from volumetric CT scans, using a novel Gaussian process Statistical Shape Model. We have demonstrated the performance of our method on a dataset of real-world patient scans, our GPDSSM model can reliably detect the presence of dysplasia. Furthermore, our GPDSSM method (accuracy 90.2\% and AUC 96.2\%) demonstrably improves on predictions based on classical angle measures such as AI and LCEA (accuracy 87.0\% and AUC 91.2\%), as well as traditional LDDMM-based computational anatomy methods (accuracy 88.0\% and AUC 92.7\%).  Our model utilises the full geometry of the acetabular surface to detect variations most correlated with dysplastic status, and is not dependent on a fixed $2$D CT slice and angle of acquisition/measurement accuracy. The speed of inference and classification algorithm can save valuable time for clinicians, removing the need for manually marking and measuring angles, as well as improving detection accuracy. Given a new patient, clinicians may use our trained GPDSSM to infer dysplastic status as follows.
\begin{enumerate}
    \item Use segmentation and isosurface extraction algorithms to segment and extract hip surface from scan.
    \item Use \cref{cup-extraction} to extract acetabular cups from obtained surfaces.
    \item Rigidly align obtained acetabula to the template of our shape model using \cref{rigid-alignment}.
    \item Fit latent point distribution to patient acetabular shape, using variational inference methodology of \cref{GPDSSM}.
    \item Use pre-trained GP classifier on latent space, to infer probability of dysplastic status.
\end{enumerate}
The whole process, given surface data, takes a couple of minutes, though some manual monitoring is required to ensure the algorithm correctly picks out the acetabular correctly. 
If the patient is dysplastic, one may further use the type of deformation heat-map analysis presented in \cref{visualisationsec}, for understanding the mode of disease variation in a patient scan, via the following steps. 
\begin{enumerate}
    \item Optimise for the closest control shape in latent space.
    \item Visualise the heat map of obtained pointwise deformation from control shape. 
\end{enumerate}
This process and the obtained deformation map will suggest the regions of the acetabulum, most likely affected by dysplasia as understood by the model. The classification and deformation heatmap may be combined with expert medical knowledge in order to personalise understanding of a patient's progress, and infer potential treatment pathways. A particularly unique aspect of this type of visualisation is that it will enable clinicians to understand the interior shape of the acetabulum. Whilst it is relatively straightforward to determine how surgical correction will increase the amount of acetabular cover, the assessment of acetabular concavity is difficult with current imaging methods. The degree of concavity is clearly important; performing an osteotomy which brings non-conforming surfaces into contact is not desirable.   The development of this procedure into an easily visualised and interpretable software package is needed for this methodology to be clinically accessible.

A limitation of the current study is the relatively small number of imaging data sets available for the modelling work. In addition, the data are from two clinical centres, which means there is a risk of bias in the sample and limited generalisability. Using more data from a wide selection of centres will improve the model utility and generalisability.

Finally, in a clinical setting, one may also utilise additional information, such as the age and sex of the patient, which is correlated with dysplastic status. This may be incorporated as additional features along with the latent-space representation, leading to a richer set of inputs for training a classifier. However, one must ensure the resulting training datasets are balanced in terms of features such as age and gender. For example, patient populations with dysplasia will tend to be skewed towards being older and/or female. Without proper rebalancing/sampling of the population, this may lead to biased classifiers that give more weight to age/gender and dominate over the latent-space embedding, which contains more subtle clinically relevant variations for classification.  

\textbf{Competing interests:} None declared.

\textbf{Funding:} Engineering and Physical Sciences Research Council (UK).

%% file: Appendix.tex
\appendix


\appendix

\section{Details of GPLVM shape model}\label{GPLVM background}

\subsection{Model specification, fitting and inference}
Here we give a compact description of the GPDSSM model specification and fitting procedure.  One may compactly describe the GPDSSM construction given an anatomical shape dataset $Y=\{S_{i}\}_{i=1}^{N}$, as follows:
\begin{enumerate}
\item Fix a \textbf{template shape} $T$, which may be chosen from the dataset (e.g. using the medoid).
\item Specify a \textbf{prior over a low-dimensional latent space} $Z=\R^{k}$; e.g., $p_{Z}=\mathcal{N}(0,I_{k})$.
\item Place a \textbf{GPLVM prior over time-dependent vector fields} $v(t,x,z)$ for $(t,x,z) \in [0,1]\times \R^{3}\times Z$. We do this in two stages. First, we place a GP prior over time-dependent `momentum' functions, 
\begin{align*}
   \bm{\alpha}(t,z) = (\alpha_{1}(t,z),\dots,\alpha_{n}(t,z)) \sim \mathcal{GP}(0,k((t,z),(t',z'))I_{nd})
\end{align*}
where $n$ is the template resolution and the process varies over latent space and time variables. Subsequently, we induce time-dependent vector fields from $\bm{\alpha}(t,z)$ as
 \begin{align*}
     v(t,x,z) = \sum_{i=1}^{n}{K_{p}(x,x_{i}(t))\alpha_{i}(t,z)} 
 \end{align*}
 where $K_{p}$ is a Gaussian spatial kernel, and where the trajectories $\bm{x}(t) = (x_{1}(t),\dots,x_{n}(t))$ satisfy the ordinary differential equation (ODE)
 \begin{align}\label{tempflow}
      \partial_{t}x_{i}(t,z) = \sum_{j=1}^{n}{K_{p}(x_{i}(t),x_{j}(t))\alpha_{j}(t,z)},\quad x_{i}(0) = x_{i},\quad i=1,\dots,n
 \end{align}
 Solutions at time $t=1$ to the above flow may be computed using numerical ODE solvers; The ODE solution at final time $t=1$ is denoted $J(\alpha,z)$.
\item Define a \textbf{likelihood for the observed shape data}, conditional on a given latent point $z$ and sampled function $\alpha(\cdot,z)$. This is done using a suitable fidelity metric $d$ on shape data as
\begin{align*}
p(S|J(\alpha,z) )\propto e^{-d(S,J(\alpha,z))^{2}}.
\end{align*}
As anatomical surface data are typically available without known correspondences, we compare shape data in the varifolds metric \cite{Younes}, which allows to make correspondence-less comparisons of shapes in a \tit{geometric} manner that accounts for the structure and connectivity of the underlying shapes. 
\end{enumerate}

The above prior and likelihood specification yields the following joint probability model:
\begin{align*}
    p(Y,\bm\alpha,U,Z)  = \prod_{i=1}^{N}{p(S_{i}\mid J(\alpha_{i},z_{i}))\,p(z_i)}\,p(\alpha|U,Z)\,p(U),
\end{align*}
where we further condition on $p(U)$ a prior distribution over \tit{inducing variables} $U$ \cite{GPLVM} in time and latent space, where $U$ is equal to $\alpha$ evaluated at $I_{U}$, a set of $m<N$ free \tit{inducing points} to be optimized in $[0,1]\times Z$. One defines $p(\bm\alpha\mid U,Z)$ through standard Gaussian conditioning formulae.

One may use a Bayesian inference procedure to infer a posterior process that best models the data under the prior regularity constraints. For speed of inference and tractable computational cost, we resort to variational posterior inference \cite{GPLVM} to find approximate posterior distributions over $\{z_{i}\}_{i=1}^{N}$ the latent representations, $U$ the inducing variable and $\alpha$ the GPLVM momenta map $\alpha(t,z)$. One does so by maximizing the following variational objective
\begin{align}\label{ELBO}
    \mathcal{L}(Y,\theta) = -\E_{q(U)q(Z)p(f|Z,U)}\bigg[\sum_{i=1}^{n}{{d(\mu_{S_{i}},\mu_{J(f_{i},z_{i})}) }^{2} }\bigg] - \mathrm{KL}(q(Z)\,||\,p(Z)) - \mathrm{KL}(q(U)\,||\,p(U)),
\end{align}
as a function of $\theta$ which contains the parameters of $q(Z)$ the latent space posterior approximation, $q(U)$ the inducing variable posterior approximation, and the hyperparameters $\theta_{h}$ of the GP and spatial kernels. The posterior for the momentum GPLVM is approximated as $p(\alpha\mid Y,U,Z)\,p(U\mid Y) \approx p(\alpha \mid U,Z)\,q(U)$. It may be shown maximizing the above as a function of $q(U),q(Z)$ approximates the true (intractable) posterior, and maximizing with respect to hyperparameters maximizes the marginal likelihood. This is done in practice by minimizing the negative of the objective function \cref{ELBO} using a stochastic variational inference approach. \cref{ELBO} contains three terms; a data term encouraging posterior fit to training shapes, and two Kullback–-Leibler (KL) divergence terms regularizing $q(U),q(Z)$.   
After fitting, given an unseen test shape $S^{*}$, one may optimize a modified variational objective $\mathcal{L}(Y\cup S^{*} ,\theta \cup \theta_{q(z^{*})})$ as a function of only $\theta_{q(z^{*})}$, containing mean and variances of $q(z^{*})$, which encode the posterior latent space distribution of the new shape. 
\subsection{Experimental configuration for SSMs}
For the experiments in \cref{Numexpsec}, we use the following model configurations for the GPDSSM (described above) and control point LDDMM model. 
\begin{enumerate}
    \item GPDSSM:  For this model, we specify Gaussian time, space and latent kernels with hyperparameters to be optimized during fitting procedure. Initial control points are sampled on the template and are fixed (not optimized). The posterior variational distributions $q(U),q(Z)$ are multivariate Gaussian with mean and covariance parameters to be optimized. The variational lower bound \eqref{ELBO} is optimized using an Stochastic Variational Inference (SVI) approach  with the Adam optimizer. The latent space dimension is fixed at size $k=20$.
    \item LDDMM shooting: We use a standard control point LDDMM atlas fitting approach presented in \cite{Younes}. For fairness of comparison we use the same choice of spatial kernel, lengthscale and control point positions as for the GPDSSM. The model fitting and reconstruction procedure results in a set of `initial' momenta vectors, one for each subject. The `latent' space representation of each subject shape in this model are obtained through PCA on the fitted set of initial momenta, with the resulting embedding lying in a subspace of dimension $k=20$ (same as for GPDSSM).
    
\end{enumerate}

\section{Alignment method}
The algorithm we use for alignment is detailed in \cref{alignmentalg} which aligns a surface $T$ to surface $S$ up to scale, rotation and translation through optimization of a metric on surfaces. In particular, the `varifolds metric' \cite{Younes} which geometrically compares triangulated surfaces through a kernel mean embedding metric, and does not require point correspondences to compute.
\begin{algorithm}[H]
\caption{Correspondence-less alignment of surface $T$ to $S$.}\label{rigid-alignment}\label{alignmentalg}
\begin{algorithmic}[1]
\State Initialise the rotation matrix, scale factor and translation vector to 
\begin{align*}
(R,c,t)=(I,1,0) \in \R^{3\times 3}\times \R \times \R^{3},
\end{align*}
and set an error tolerance $\delta>0$.
\State Define objective function $E({(R,c,t)}):= \norm{\mu_{S}- \mu_{cR(T) + t} }_{V^{*}}^{2}$, using the varifolds metric $\norm{\cdot}_{V}$.
\While{$E({(R,c,t)})>\delta$}{
\State Compute gradients $\nabla_{(R,c,t)} E({(R,c,t)})$ using automatic differentiation.
\State Update parameters $(R,c,t)$ via step of chosen gradient based optimizer.

\EndWhile
}
\end{algorithmic}
\end{algorithm}
In practice, we parametrize rotation matrices in \cref{alignmentalg} through a quaternion representation commonly used in shape-modelling applications; e.g., see the appendix of \cite{Younes} for details on this paramerization.

\section{Shape modelling and classification}
Our proposed method is based on constructing a Statistical Shape Model (SSM) for the geometry of the acetabulum. We build our shape model by combining the Gaussian Process Latent Variable Model (GPLVM) \cite{GPLVM} with the Large Deformation Diffeomorphic Metric Mapping \cite{Younes} (LDDMM) theory. This model is henceforth referred to as the Gaussian Process Diffeomorphic Statistical Shape Model (GPDSSM); a generative statistical model of shape with topological guarantees that yields a low-dimensional latent space embedding of shapes. Given a dataset of (left and right) acetabular surfaces, our shape model will both learn the underlying shape variation, and automatically infer low-dimensional latent vector representations representing the acetabular geometry of each subject. The latent embeddings will be sensitive to variations in shape due to dysplasia and implicitly contain information about the correlation between shape and dysplastic status. This information can be exploited by fitting a classification model for dysplastic status on the low-dimensional latent space. Our model may be fit from a dataset of triangulated surfaces using Bayesian inference techniques, providing posterior distributions over latent embeddings of shapes that account for model and data uncertainty. Furthermore, fitting does not rely on having point-to-point correspondences in the given shape data, making it easily adapted to real-world shape modelling applications where one often lacks parametric correspondence.  In the remainder of this section we describe our choice of SSM and its construction in simple terms.

\subsection{Gaussian Process Latent Variable Model}\label{GPLVM}

At the core of our shape model is a GPLVM \cite{GPLVM}, a statistical generative model with automatic nonlinear dimensionality reduction. A GPLVM may be expressed simply as
\begin{align}\label{simplegaussian}
    y = f(z) + \varepsilon,
\end{align}
where $y$ is observed data, and $f: Z \longrightarrow Y$ is a randomized mapping from latent space ${Z}$, to ${Y}$ the data space. The noise in the data is typically modelled by a Gaussian random variable. Importantly, $f$ is a \tit{Gaussian} process \cite{gpml}, meaning the evaluations of $f$ at any finite set of latent points has a joint Gaussian distribution with mean and covariance determined by a mean function $m_{\theta}(z)$ and covariance kernel $k_{\theta}(z,z')$ with learnable hyperparameters $\theta$. This is often denoted compactly as
\begin{align}\label{firstGPprior}
    f(z) \sim \mathcal{GP}(m_{\theta}(z),k_{\theta}(z,z')).
\end{align}
The regularity properties of $m,k$ determine the sample properties of $f$.

 As opposed to the regression setting, in generative modelling the input variables $z$ are unknown, and we only have access to $y$ data. As such, given a dataset $\mathcal{D}=\{y_{i}\}_{i=1}^{n}$, one fits the model by simultaneously inferring a suitable latent variable $z_{i}$ for each data point $y_{i}$, as well as a Bayesian posterior distribution over mapping $f$ in order to best represent the data generating process. This is often done using a variational inference framework, by maximizing a lower bound on the marginal log-likelihood of \cref{simplegaussian,firstGPprior} as a function of variational posterior distributions over $f,z_{i},\theta$.
 
 After fitting, one may generate new data with a similar variation to the dataset by sampling the fitted posterior distribution over $f$ at new $z$ locations. An attractive feature of the GPLVM is that under suitable constraints, the GPLVM automatically places data with similar features close together in latent space and dissimilar data in different regions in the latent space. This property is useful for down-line tasks such as classification and clustering of data, as such tasks may be performed in the low-dimensional latent space, which has `learned' the underlying relationships/structure in the data. 
\subsection{LDDMM}\label{LDDMM}
To adapt the GPLVM to shape data, we combine it with the LDDMM framework, which is widely used in computational anatomy \cite{Tangentstats,Younes} for modelling and understanding variations in anatomical shapes due to degenerative diseases. In LDDMM, anatomical objects $S$ are modelled as deformations of a template anatomy $T$ under the action of a \tit{diffeomorphism} $\varphi$ on the underlying domain, 
\begin{align}\label{basicLDDMM}
S =  \varphi(T) + \varepsilon.
\end{align}
In the notation of \cref{basicLDDMM}, the diffeomorphism $\varphi$ \tit{deforms} the template $T$ to produce a new anatomy/shape $\varphi(T)$. We denote by $\varepsilon$ a suitable noise distribution on the shape data. Diffeomorphisms are continuously differentiable maps from the underlying domain to itself with a continuously differentiable inverse. Such mappings are known \cite{Younes} to preserve topological features of the shapes on which they act while being flexible enough to describe a rich class of shape variations. This makes diffeomorphic mappings a natural choice for statistical shape modelling, especially when dealing with a class of shapes that exhibit complex variations while sharing the same implicit topological constraints.

Diffeomorphisms may be parametrized in practice as \textit{flows} $\varphi^{v}$ of time-varying deformation fields $v(t,x)$ that are sufficiently regular. In this notation, $v(t,x)$ is the infinitesimal deformation vector of the point $x$ in the domain at time $t$. The flow is computed for each point $x$ by effectively accumulating these deformations over time through integration. Given $v$, in practice one may compute its flow mapping $\varphi^{v}$ using numerical ordinary differential equation (ODE) solvers. A generative model for time dependent deformation fields $v_{\theta}(z,t,x)$ thus induces a model for shapes as
\begin{align}\label{parametrizedLDDMM}
    S = \varphi^{v_{\theta}(z,\cdot,\cdot)}(T) + \varepsilon.
\end{align}
In \cref{parametrizedLDDMM}, a generative model for fields $v_{\theta}(z,t,x)$ induces a generative model over flows $\varphi^{v_{\theta}(z)}(T)$, which deform the template to produce new shapes for each latent variable $z$. One may learn $v_{\theta}$ from shape data, allowing one to effectively learn diffeomorphic variation in a shape dataset with consistent topology. 

\subsection{Gaussian Process Diffeomorphic Statistical Shape Model (GPDSSM)}\label{GPDSSM}
 In this work, we parametrize $v_{\theta}(z,\cdot,\cdot)$ by combining the probabilistic framework of the GPLVM with the expressivity and topological guarantees of LDDMM. In \cref{parametrizedLDDMM} we place a prior over deformation fields using a GPLVM as
\begin{align}\label{fieldprior}
    v_{\theta}(z,t,x) \sim \mathcal{GP}(0,k_{\theta}((z,t,x),(z',t',x'))),
\end{align}

Given a dataset of shapes $\mathcal{S} = \{S_{1},\dots,S_{N}\}$, the resulting prior model from combining equations \cref{parametrizedLDDMM,fieldprior} is expressed compactly as
\begin{align}\label{GPDSSM-model}
  S_{i} = \varphi^{v_{\theta}(z_{i},\cdot,\cdot)}(T) + \varepsilon_{i},\quad i=1,\dots,N.
\end{align}
In this model, each shape $S_{i}$ in the dataset is modelled as a noisy deformation of the template $T$. Each deformation is generated as the flow of a random time-dependent vector field ${v_{\theta}(z_{i},\cdot,\cdot)}$, sampled from the prior GPLVM \cref{fieldprior} at fixed latent point $z_{i}$. One may use a similar Bayesian variational posterior inference framework as in the standard GPLVM to fit this model. This allows one to automatically infer a posterior distribution over the deformation fields $v_{\theta}(z,t,x)$ (and it's hyperparameters), as well as low-dimensional latent embeddings $\{z_{i}\}_{i=1}^{N}$ representing each subject in the anatomical shape dataset $\{S_{i}\}_{i=1}^{N}$.

\subsection{Classification of shape data}\label{classificatintheorysec}
We now discuss how to use the statistical shape model presented in the previous section to classify subjects for hip dysplasia. Henceforth, we denote the shape model presented in the previous section as $\mathcal{M}$. We assume one is given a labelled set of aligned acetabular surfaces
\begin{align*}
\mathcal{S} = \{(S_{1},y_{1}) ,\dots,(S_{N},y_{N})\},
\end{align*}
 obtained and pre-processed with the methodology presented in \cref{datasec}, where $S_{i}$ denote the surface of the $i$'th subject anatomy and $y_{i} \in \{0,1\}$ the corresponding label; $1$ for dysplastics and $0$ for controls. After fitting $\mathcal{M}$ to the shapes $S_{i}$, this yields a set 
\begin{align*}
\mathcal{Y} = \{(S_{1},z_{1}),\dots,(S_{N},z_{N})\}
\end{align*} 
of surfaces $S_{i}$ and associated low-dimensional latent vectors $z_{i} \in Z$ learned by the model for each shape in the dataset. At this stage, $\mathcal{M}$ has only been fitted to the surfaces $S_{i}$ and has no access to the labels $y_{i}$. However, natural correlations exist between between the geometry of the acetabulum and dysplasia status. As such, one would expect shapes with similar (dysplastic) shape variations will cluster together in the low-dimensional latent-space embedding. One may build on this intuition to train a classification algorithm $\mathcal{A}$ on the set
\begin{align*}
    \mathcal{T} = \{(z_{1},y_{1}),\dots,(z_{N},y_{N})\}
\end{align*}
to exploit the correlations in the latent-space representation, shape and dysplastic status. In practice, we choose $\mathcal{A}$ to be a probabilistic classifier which outputs a score in the range $[0,1]$ indicating the likelihood of the presence of dysplasia. The decision boundaries of the classifier on $Z$ will delineate which regions correspond to (high probability) dysplastic shapes and which regions to control shapes.

After fitting both $\mathcal{M}$ and $\mathcal{A}$, given a new subject scan, clinicians may use our trained GPDSSM to infer dysplastic status as follows.
\begin{enumerate}
    \item Segment and extract pelvis socket surface automatically from scan.
    \item Use \cref{cup-extraction} to extract acetabular surface, rigidly aligned to template of shape model using \cref{rigid-alignment}.
    \item Fit latent point $z \in Z$ to subject in latent space of $\mathcal{M}$, using methodology of \cref{GPDSSM}.
    \item Use pre-trained classifier $\mathcal{A}$ on latent space, to infer probability of dysplastic status.
\end{enumerate}
The last two steps are computationally cheap and fast to evaluate; the first reduces to a low-dimensional optimization problem over $Z$ with model parameters fixed, and the second requires to evaluate our pre-trained classifier $\mathcal{A}$ on a single point. Therefore, once training of $\mathcal{M},\mathcal{A}$ is complete, the above yields a simple procedure for screening new patients for dysplasia, given patient hip scans/surfaces. 